\documentclass[]{fairmeta}

\usepackage{xcolor}
\usepackage{amsmath}
\usepackage{amssymb}
\usepackage{booktabs}
\usepackage{makecell}     
\usepackage{placeins}     

\newcommand{\method}{MobileWAM}
\newcommand{\cof}{CoF}
\newcommand{\bs}[1]{\boldsymbol{#1}}

\title{MobileWAM: Bridging World Action Models to Mobile Manipulation with Chain-of-Foresight}

\author[1,2,\dagger]{Zehua Fan}
\author[3,\dagger]{Junjie He}
\author[3,\dagger]{Wenxuan Song}
\author[4]{Xi Wang}
\author[5]{Wenqi Lyu}
\author[6]{Linge Zhao}
\author[3]{Fuhao Li}
\author[7]{Zihan You}
\author[8]{Yifei Yang}
\author[9]{Kaiming Xu}
\author[10]{Qi Jiang}
\author[10]{Yue Jiang}
\author[3]{Haoang Li}
\author[11,*]{Cheng Chi}
\author[2]{Feng Gao}
\author[10,*]{Bailin Li}
\author[1,*]{Yan Wang}

\contribution[\dagger]{Equal contribution}
\contribution[*]{Corresponding authors}

\affiliation[1]{Institute for AI Industry Research (AIR), Tsinghua University}
\affiliation[2]{Shanghai Jiao Tong University}
\affiliation[3]{The Hong Kong University of Science and Technology (Guangzhou)}
\affiliation[4]{AIR Wuxi Innovation Center, Tsinghua University}
\affiliation[5]{The University of Adelaide}
\affiliation[6]{Wuhan University}
\affiliation[7]{Southeast University}
\affiliation[8]{Beijing Jiaotong University}
\affiliation[9]{Fudan University}
\affiliation[10]{Li Auto}
\affiliation[11]{School of Information, Renmin University of China}

\abstract{
World action models (WAMs) built on video generation backbones are a rising recipe for robot learning, yet remain confined to tabletop manipulation. Mobile manipulation demands simultaneous locomotion and whole-body manipulation amid scene-scale dynamics, yet is still dominated by dynamics-blind visual encoders with hand-crafted coordination. We bridge this gap with \method{}, a mixture-of-transformers architecture that fuses a pretrained video diffusion transformer with a lightweight action expert through layerwise joint attention, translating internet-scale motion priors into whole-body control. To reconcile the heterogeneous dynamics of moving and manipulating, each feed-forward layer of the action expert becomes a three-expert mixture of shared, locomotion, and manipulation experts, softly routed by the motion intent in the action tokens. To densify supervision, we further propose Chain-of-Foresight (\cof{}): intermediate representations sequentially predict a chain of future latent chunks, each step conditioned on its predecessor. \cof{} pairs naturally with our decoupled video--action denoising scheme. At deployment, the WAM serves as a pure current-frame encoder; foresight acts only through gradients, so at inference the foresight chain and video generation are discarded, leaving only policy-level cost. \method{} surpasses state-of-the-art mobile manipulation policies on ManiSkill-HAB and fine-tunes to a real ARX Lift2 mobile manipulator across diverse tasks with strong generalization. Code will be released upon acceptance.
}

\date{\today}

\begin{document}

\maketitle

\begin{figure*}[t]
\centering
\includegraphics[width=0.97\textwidth]{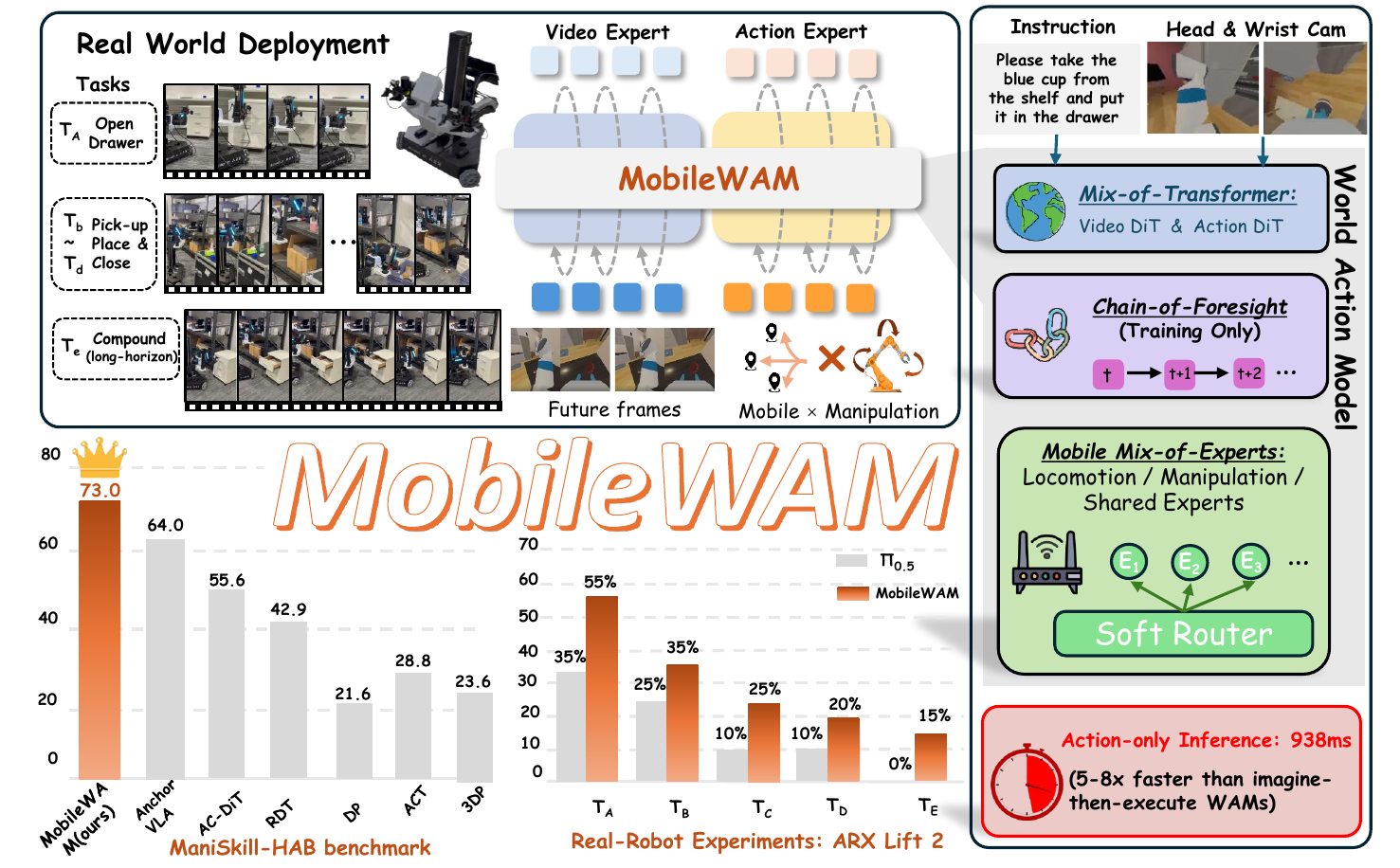}
\caption{\textbf{\method{} at a glance.} \emph{Top-left:} five real-robot tasks of increasing horizon ($T_a$--$T_e$) on an ARX Lift2. \emph{Top-center/right:} Mixture-of-Transformers fuses video and action experts; Chain-of-Foresight (training only) chains future-latent predictions via an RNN-style belief; Mobile MoE routes locomotion, manipulation, and shared experts via a soft router. \emph{Bottom-left:} ManiSkill-HAB mean S.R.\ 73.0\%, leading all listed baselines. \emph{Bottom-right:} real-robot success rates vs.\ $\pi_{0.5}$ across $T_a$--$T_e$; action-only inference at 938\,ms is 5--8$\times$ faster than generate-then-act WAMs.}
\label{fig:teaser}
\end{figure*}

\section{Introduction}

Is mobile manipulation merely tabletop manipulation with a mobile base, or ``navigation plus manipulation'' stapled together? We argue no: coupling locomotion with manipulation is multiplicative rather than additive. Three gaps separate the two settings. \emph{Viewpoint}: tabletop cameras are fixed; mobile robots' cameras sweep across places, demanding perception consistent under relentless ego-motion. \emph{Multimodality}: with the base free, many base--arm paths reach the same goal, making the action distribution far more multimodal~\citep{lim2026anchorvla}. \emph{Causality}: an early navigation error silently dooms a later grasp. Mobile manipulation is long-horizon not by frame count, but by causal depth and spatial variation.

Tabletop tasks often reduce to vision-to-action mappings where short-horizon VLAs excel~\citep{brohan2023rt2, kim2024openvla}. These shallow mappings evaporate in mobile manipulation, where representations must encode how the world \emph{will} evolve.

Video-generative diffusion transformers address this need: pretrained on internet-scale video, they internalize physical priors and synthesize coherent futures, fueling world action models (WAMs)~\citep{bi2026motus, li2026lingbotva, yuan2026fastwam}. Yet successes concentrate on fixed-base manipulation; mobile manipulation still relies on dynamics-blind encoders: point-cloud fusion with two-stage conditioning~\citep{chen2025acdit}, or decoders supervised with privileged GT segmentation~\citep{tu2026sgvla, lin2026wem}. The natural marriage has yet to be consummated.

We take an early step with \method{} (Figure~\ref{fig:teaser}), a WAM systematically adapted to mobile manipulation. It grafts a lightweight action expert onto a pretrained video diffusion transformer via layerwise joint attention, mixture-of-transformers style~\citep{liang2025mot}, steeping whole-body action denoising in the backbone's motion priors at every layer. Two designs tailor it further. First, locomotion and manipulation are distinct regimes; forcing one pathway to serve both causes destructive interference. Our \emph{mobile mixture-of-experts} routes each feed-forward layer across three experts (shared, locomotion, manipulation) softly routed by motion intent. Second, inspired by next-latent prediction~\citep{teoh2025nextlat} in LLMs, we propose \emph{Chain-of-Foresight} (\cof{}) that embeds a latent dynamics model: an RNN-style chain where step $k$ denoises the $k$-th future latent and passes its belief to step $k{+}1$ with hidden states, which carry the world's physical state forward. Following the decoupled denoising paradigm~\citep{yuan2026fastwam}, deployed \method{} acts as a \emph{current-frame encoder} that caches one backbone pass and denoises actions against it. \cof{} is deleted at inference, yielding $5\times$--$8\times$ speedups over imagine-then-execute WAMs.

\method{} surpasses state-of-the-art on ManiSkill-HAB~\citep{shukla2025mshab} with RGB-only inputs and one-stage training, and transfers to a real ARX Lift2, beating $\pi_{0.5}$~\citep{black2025pi05} with margins that grow with task horizon.

Our contributions are fourfold:

{\setlength{\parskip}{0pt}%
\noindent\textbullet~\textbf{Bridging WAMs to mobile manipulation.} To our knowledge, we present the first systematic adaptation of video-generation WAMs to whole-body mobile manipulation, achieving state-of-the-art ManiSkill-HAB results with RGB-only inputs and one-stage training.

\noindent\textbullet~\textbf{Chain-of-Foresight.} A recurrent, serially chained future-latent prediction objective that densifies temporal supervision and strengthens long-range spatio-temporal and causal consistency, at zero inference cost.

\noindent\textbullet~\textbf{Mobile MoE.} A three-expert mixture (shared, locomotion, manipulation) inside the action expert, routed by motion intent, reconciling the heterogeneous dynamics of moving and manipulating.

\noindent\textbullet~\textbf{Real-robot deployment.} \method{} fine-tunes to the ARX Lift2 and consistently outperforms $\pi_{0.5}$ across five household tasks, with the largest gains on the longest horizons.
}

\section{Related Work}

\subsection{World Action Models}
World action models couple a generative observation model with an action head: imagining motion consequences yields physically grounded representations. Early systems were two-stage: UniPi inverts a generated video into actions~\citep{du2023unipi}; VPP and Seer condition policies on predicted states~\citep{hu2025vpp, tian2025seer}. Recent WAMs collapse stages into end-to-end pretrained diffusion transformers: Motus~\citep{bi2026motus}, LingBot-VA~\citep{li2026lingbotva}, Fast-WAM~\citep{yuan2026fastwam}, GigaWorld-Policy~\citep{gigaworld2026policy}; ImageWAM even finds an image-editing backbone sufficient for tabletop WAMs~\citep{zhang2026imagewam}, suggesting fixed-base manipulation barely needs long-horizon video prediction. S-VAM pushes the inference--foresight trade-off further by self-distilling multi-step generative priors into a single forward pass~\citep{yan2026svam}. Yet all address fixed-base tasks, blunting their core advantage. \method{} adopts this line's mixture-of-transformers fusion and decoupled denoising, but extends it to where the camera rides on the robot base.

\subsection{Mobile Manipulation}
Mobile manipulation research splits into two camps. The \emph{modular} camp decouples navigation from manipulation via trajectory optimization~\citep{wu2024remani, yan2025m2diffuser}, base-pose optimization~\citep{yang2025mobipi, wu2025moto, wu2025momanipvla}, or LLM/VLM planners~\citep{wang2025ialp, chen2025owmm}, at the price of hand-designed interfaces that leak errors across stages. The \emph{end-to-end} camp learns whole-body control directly~\citep{fu2024mobilealoha, jiang2025brs, li2025momagen, zhu2026emma}: dense policies~\citep{su2025dspv2}, anchored diffusion~\citep{lim2026anchorvla}, memory-augmented VLAs~\citep{lin2025echovla}, reasoning-augmented VLAs that inject visual-linguistic chains of thought~\citep{zhong2026dualcotvla}, and $\pi_0$-family models dominating BEHAVIOR~\citep{black2024pi0, black2025pi05}. Closest to ours: AC-DiT requires 3D point clouds and two-stage training~\citep{chen2025acdit}; SG-VLA uses auxiliary decoders with privileged GT segmentation~\citep{tu2026sgvla}; WEM separates world and ego streams but presupposes that separation~\citep{lin2026wem}; ABot-M0.5 explores WAMs with latent actions~\citep{chen2026abot}. \method{} needs none of the above. Its advantage runs deeper than a leaner recipe: the consistency others engineer explicitly is inherited from video pretraining and reinforced from within. \cof{} turns intermediate representations into an RNN-style latent dynamics model that directly supervises physical regularity, while the mobile MoE lets locomotion and manipulation experts specialize without severing the shared structure that binds them.

\subsection{Future Prediction as Auxiliary Supervision}
A complementary line supervises \emph{representations} with the future rather than generating pixels: latent-dynamics planning over frozen features~\citep{zhou2025dinowm}, foresight-conditioned inverse dynamics~\citep{tian2025seer}, and reconstruction-based perception for VLAs~\citep{song2026reconvla}. Most related, the Belief State Transformer and Next-Latent Prediction show that predicting one's \emph{own future latent state} injects a recurrent inductive bias, compressing history into belief states~\citep{hu2025bst, teoh2025nextlat}. \cof{} transplants this into WAMs with two twists: targets are \emph{future video latents} grounded in physical scene evolution, and the chain is serially unrolled so each step conditions on its predecessor's belief. The chain is never consulted at inference; it exists solely to shape the representations the action expert reads.

\begin{figure*}[t]
\centering
\includegraphics[width=0.96\textwidth]{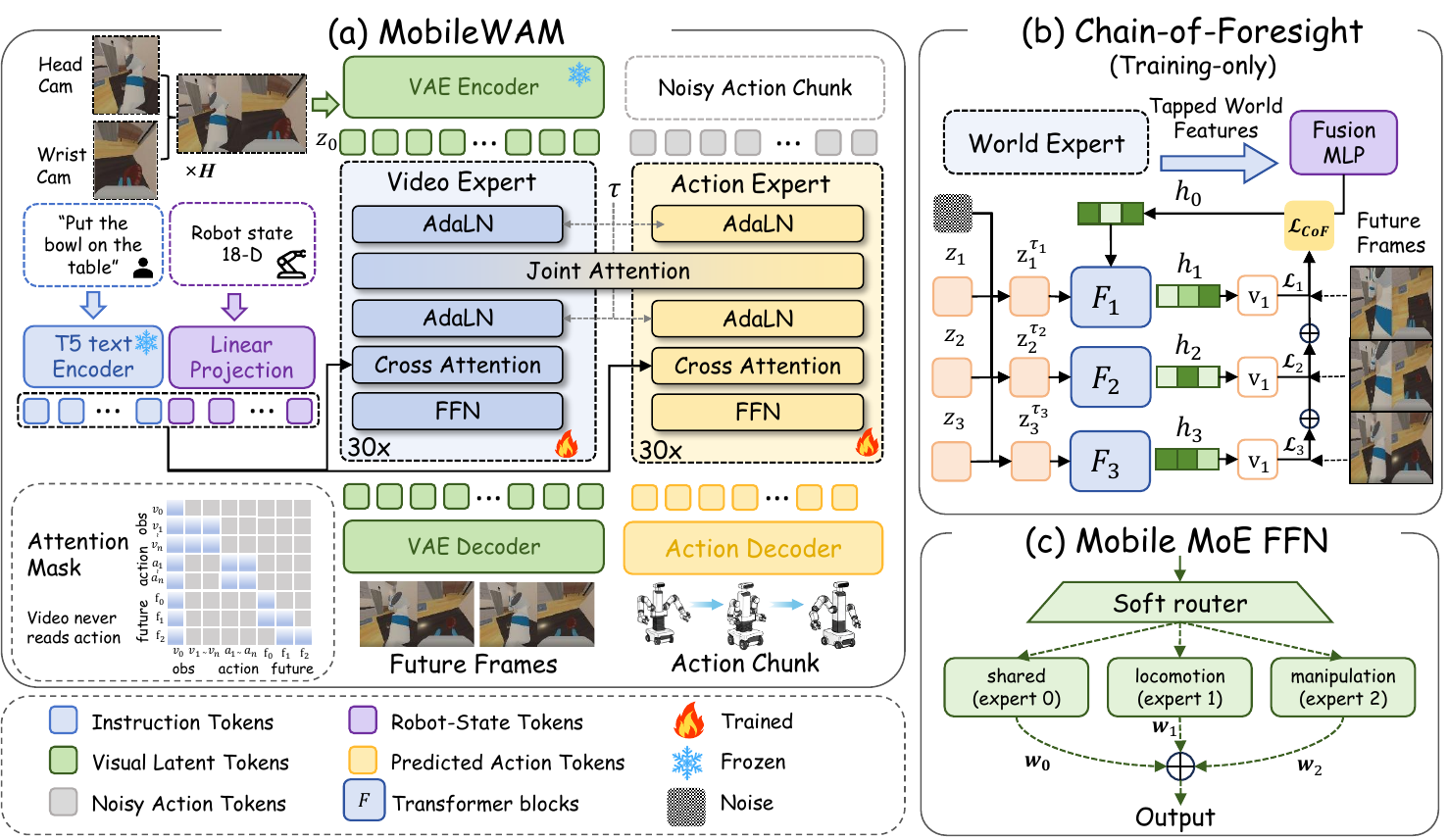}
\caption{\textbf{\method{} architecture.}
\textbf{(a) Main model.} The world and action experts fuse via asymmetric layerwise joint attention (actions read visuals; visuals never read actions); language and proprioception are injected via per-block cross-attention; every feed-forward layer of the action expert is a three-expert mobile MoE (shared $\oplus$ locomotion $\oplus$ manipulation) softly routed by motion-intent embeddings.
\textbf{(b) Chain-of-Foresight.} Four uniformly spaced backbone layers fuse into belief $\bs{h}_0$; depth-specific $F_1,F_2,F_3$ (separate weights, init.\ from later backbone blocks) chain over $f_0,f_1,f_2$, each attending to the previous belief and current observation. Weights $w_1{>}w_2{>}w_3$ discount farther futures. The \cof{} branch is discarded at inference.
\textbf{(c) Attention mask.} Token groups: $v_0$ (current observation, clean latent); $v_1{\sim}v_n$ (future video latents, noisy, bidirectional); $a_1{\sim}a_n$ (noisy action chunks); $f_0{\sim}f_{K-1}$ (foresight latents, training only). $v_1{\sim}v_n$ attend to $v_0$ and each other; actions attend to $v_0$ only; each foresight step $f_k$ ($k{\ge}1$) attends to $v_0$ and $f_{k-1}$, while $f_0$ attends to $v_0$ and itself. Video and action denoising are fully decoupled, enabling current-frame-encoder inference.}
\label{fig:arch}
\end{figure*}

\section{Method}

\subsection{Overview}
At time $t$ the robot receives an observation $o_t{=}\{I^{\mathrm{head}}_t, I^{\mathrm{wrist}}_t\}$ from head and wrist RGB cameras, proprioceptive state $\bs{s}_t\in\mathbb{R}^{d_s}$ (joint positions, velocities, and base odometry), and language instruction $\ell$. The policy outputs a whole-body action chunk $\bs{a}_{t:t+H}{=}(\bs{a}_t,\dots,\bs{a}_{t+H-1})$ with each $\bs{a}\in\mathbb{R}^{d_a}$; $d_a{=}13$ dimensions jointly command the arm (7 joints), gripper, head pan/tilt, torso lift, and base linear/angular velocities. During training the model also observes a future video segment $V_{t:t+T}$ of horizon $T$ and learns
\begin{equation}
p_\theta\!\left(\bs{a}_{t:t+H},\, V_{t:t+T} \mid o_t, \bs{s}_t, \ell\right);
\end{equation}
at deployment only the action marginal is sampled. Both modalities are trained with flow matching, detailed below.

\method{} (Figure~\ref{fig:arch}) couples a large \emph{world expert} (pretrained video diffusion transformer) with a lightweight \emph{action expert} via layerwise joint attention. An asymmetric attention mask decouples video and action denoising so that deployment can drop the video branch without changing the current-frame features the policy reads. Chain-of-Foresight (\cof{}) provides a training-only recurrent supervision signal at zero inference cost.

\subsection{World Expert}
The backbone is a pretrained text-and-image-to-video diffusion transformer~\citep{wan2025wan, peebles2023dit} ($L{=}30$ blocks, width 3072). A frozen 3D VAE compresses video $16{\times}$ spatially and $4{\times}$ temporally; one latent tick covers one action chunk ($H{=}4$), aligning visual and motor clocks. Both camera views are composited side by side, enforcing cross-view consistency via spatial attention. The current frame enters as a clean latent (timestep zero); subsequent frames carry noise, so the backbone denoises the future \emph{given} the present. Language $\ell$ is encoded by a frozen T5-family encoder~\citep{raffel2020t5} and injected via per-block cross-attention.

\subsection{Action Expert and Mobile MoE}
A lightweight action-denoising transformer (width 1024) runs in lockstep: at each of the 30 layers, its action tokens join the backbone's visual tokens in shared self-attention, mixture-of-transformers style~\citep{liang2025mot, bi2026motus}, with separate projections but a concatenated sequence. Noisy action chunks are linearly embedded as action tokens; $\bs{s}_t$ is appended to the text context for cross-attention.

Locomotion and manipulation are distinct motion regimes: base commands navigate through free space; arm and gripper commands resolve precise contact. Forcing a single pathway to handle both causes destructive interference: updates that sharpen grasping compete with those shaping approach trajectories. We replace every feed-forward layer of the action expert with a three-expert mixture: a \emph{shared} expert for regime-agnostic structure, plus dedicated \emph{locomotion} and \emph{manipulation} experts~\citep{shazeer2017moe}. A router reads the mean-pooled noisy action embedding and produces soft weights via temperature-scaled softmax. Experts are cloned from the dense layer; the router is zero-initialized, so specialization emerges only where data demands it. No load-balancing loss is needed. The design echoes the disentanglement in~\citet{chen2026abot}, but as a drop-in, single-stage module.

\subsection{Chain-of-Foresight}
A single prediction chunk asks ``what happens next?'' Mobile manipulation demands ``and then? and then?'' \cof{} converts this into an auxiliary objective (Figure~\ref{fig:arch}b).

Let $\bs{H}^{(l)}$ be the backbone's hidden states at layer $l$ for current-observation tokens. We tap four uniformly spaced layers $\{4,12,20,30\}$ and fuse them via a two-layer MLP $g$:
\begin{equation}
\bs{h}_0 = g\!\left(\left[\bs{H}^{(4)};\,\bs{H}^{(12)};\,\bs{H}^{(20)};\,\bs{H}^{(30)}\right]\right).
\end{equation}
Shallow layers carry geometry; deep layers carry semantics; either extreme or all-30 concatenation degrades the chain (Table~\ref{tab:layers}).

Depth-specific modules $F_1,\dots,F_K$ (separate weights, initialized from later backbone blocks) are chained over future latents $\bs{z}_1,\dots,\bs{z}_K$ ($f_0,\dots,f_{K-1}$ in Figure~\ref{fig:arch}). At step $k$, $F_k$ takes belief $\bs{h}_{k-1}$ and noised target $\bs{z}_k^{\tau_k}$, cross-attends to $(\ell, \bs{s}_t, o_t)$ under a block-causal mask, and outputs a velocity estimate and foresight belief:
\begin{equation}
\left(\hat{\bs{v}}_k,\; \bs{h}_k\right) = F_k\!\left(\bs{h}_{k-1},\, \bs{z}_k^{\tau_k};\; \ell, \bs{s}_t, o_t\right), \quad k=1,\dots,K.
\end{equation}
The belief is the sole conduit between steps, forcing $\bs{h}_k$ to summarize the evolving world: the latent analogue of belief-state learning~\citep{hu2025bst, teoh2025nextlat}.

\subsection{Attention Mask and CoF Interaction}
The joint attention mask is asymmetric (Figure~\ref{fig:arch}): current tokens attend only among themselves; future video tokens attend to all visual tokens; action tokens attend to current tokens only; no video token sees any action token. This ensures the current-observation representation is identical at training and deployment.

At inference, the WAM becomes a \emph{current-frame encoder}: per-layer KV pairs are cached from one backbone pass; action tokens are iteratively denoised against this cache without instantiating any future frame~\citep{yuan2026fastwam}. \cof{} is fully compatible: its gradients flow from each foresight step through the fusion MLP into the tapped backbone layers, and from there into the joint attention that action tokens read, pressuring the backbone to encode scene dynamics in its current-observation representation. The foresight module and fusion MLP are deleted at inference, adding zero parameters or FLOPs.

\subsection{Losses}
\label{sec:losses}
Both video and action are trained with flow matching~\citep{lipman2023flow}. For a clean sample $\bs{x}$ (video latent or action chunk), a noisy interpolant $\bs{x}^\tau{=}(1{-}\tau)\bs{x}{+}\tau\bs{\varepsilon}$, $\bs{\varepsilon}{\sim}\mathcal{N}(\bs{0},\bs{I})$, is formed at noise level $\tau\in[0,1]$; the network predicts velocity field $\hat{\bs{v}}$ approximating $\bs{v}{=}\bs{\varepsilon}{-}\bs{x}$:
\begin{equation}
\mathcal{L}_{\mathrm{v}} = \mathbb{E}\,\|\hat{\bs{v}}_{\mathrm{vid}}-\bs{v}_{\mathrm{vid}}\|^2,\quad
\mathcal{L}_{\mathrm{a}} = \mathbb{E}\,\|\hat{\bs{v}}_{\mathrm{act}}-\bs{v}_{\mathrm{act}}\|^2,
\end{equation}
with independent, shift-skewed timestep sampling so either branch can be denoised alone. Sharing one formalism lets action tokens attend to visual tokens under the same noise semantics, making layerwise fusion coherent.

\cof{} contributes a depth-decayed term, discounting farther futures:
\begin{equation}
\mathcal{L}_{\mathrm{\cof{}}} = \sum_{k=1}^{K} w_k\, \mathbb{E}\,\|\hat{\bs{v}}_k-\bs{v}_k\|^2,\quad w_1>\cdots>w_K.
\end{equation}
The total objective is $\mathcal{L} = \mathcal{L}_{\mathrm{v}} + \mathcal{L}_{\mathrm{a}} + \lambda\,\mathcal{L}_{\mathrm{\cof{}}}$. Specific weight values are given in the Appendix.

\section{Experiments}

We evaluate \method{} on the ManiSkill-HAB benchmark, ablate every design decision along the evolution path that led to \cof{} and the mobile MoE, measure deployment latency against imagine-then-execute WAMs, and validate transfer to a real mobile manipulator.

\subsection{Setup}
\paragraph{Benchmark.}
ManiSkill-HAB~\citep{shukla2025mshab} is a GPU-parallelized, low-level-control re-implementation of the Home Assistant Benchmark~\citep{szot2021habitat2} in ManiSkill3~\citep{tao2024maniskill3}, featuring a Fetch robot performing home-scale rearrangement. We adopt the seven subtask--object--scene combinations of the \emph{SetTable} suite: picking an apple from the fridge, placing the apple on the table, opening the fridge door, picking a bowl from the counter, placing the bowl on the table, opening the counter drawer, and closing the counter drawer. Success is measured by closed-loop rollout: an episode succeeds only if the subtask's completion predicate holds at termination. We report mean and standard deviation of success rate (S.R.) over three independent evaluation runs.

\paragraph{Training.}
For each subtask, we use 1{,}000 filtered demonstration trajectories (100 held out for validation) from the benchmark's data-generation pipeline. \method{} is trained in a \emph{single} stage with AdamW (learning rate $10^{-5}$, cosine schedule, bf16), a global batch size of 256, and full fine-tuning of both experts; the VAE and text encoder stay frozen. Loss weights are $\lambda_{\mathrm{v}}{=}\lambda_{\mathrm{a}}{=}1$ and $\lambda{=}0.1$ with depth decay $\bs{w}=(0.4, 0.2, 0.1)$, $K{=}3$. Observations are $384{\times}640$ composited RGB frames; no depth, no point clouds, no privileged states. At inference, we denoise actions in 20 flow-matching steps and execute the full 4-step chunk before replanning. Ablations (Tables~\ref{tab:variants}--\ref{tab:moedesign}) use a reduced 5{,}000-step budget to expose trends economically, so their absolute numbers are lower. Further implementation details are in the Appendix.

\begin{table*}[t]
\centering
\small
\setlength{\tabcolsep}{8pt}
\renewcommand{\arraystretch}{1.15}
\caption{\textbf{Main results on ManiSkill-HAB (SetTable).} Success rate (\%, mean$\pm$std over three evaluation runs) on the seven subtask combinations. \method{} uses RGB-only inputs and a single training stage. \textbf{Bold} marks the best and \underline{underline} the second best per row (i.e., per subtask across methods); ``--'' denotes not reported. Baselines: ACT~\citep{zhao2023act}; DP~\citep{chi2023diffusionpolicy}; DP3~\citep{ze2024dp3}; RDT~\citep{liu2025rdt}; AC-DiT~\citep{chen2025acdit}; AnchorVLA~\citep{lim2026anchorvla}.}
\label{tab:main}
\begin{tabular}{lccccccc}
\toprule
\textbf{Subtask} & \textbf{ACT} & \textbf{DP} & \textbf{DP3} & \textbf{RDT} & \textbf{AC-DiT} & \textbf{AnchorVLA} & \textbf{\method{}} \\
\midrule
Pick Apple  & 28.0$\pm$2.2 & 21.3$\pm$3.3 & 0.0$\pm$0.0  & 12.0$\pm$11.3 & \underline{33.3$\pm$1.9} & 22.7$\pm$0.9 & \textbf{46.0$\pm$0.8} \\
Place Apple & 8.7$\pm$3.3  & 28.0$\pm$8.0 & 31.0$\pm$0.8 & 32.0$\pm$5.7  & 33.3$\pm$9.4              & \textbf{64.3$\pm$0.8} & \underline{63.7$\pm$3.2} \\
Open Fridge & 2.0$\pm$2.2  & 7.3$\pm$5.8  & 0.0$\pm$0.0  & 82.7$\pm$10.5 & \underline{90.7$\pm$5.0} & 88.9$\pm$0.8 & \textbf{99.3$\pm$0.5} \\
Pick Bowl   & 28.0$\pm$2.4 & 20.7$\pm$3.3 & 20.0$\pm$2.4 & 10.7$\pm$6.8  & 36.0$\pm$6.5 & \underline{44.5$\pm$0.8} & \textbf{46.0$\pm$2.6} \\
Place Bowl  & 13.0$\pm$0.8 & \textbf{69.3$\pm$3.3} & 32.0$\pm$0.8 & 18.7$\pm$5.0 & 17.3$\pm$6.8 & 63.8$\pm$2.0 & \underline{64.7$\pm$1.2} \\
Open Drawer & 0.0$\pm$0.0  & 0.0$\pm$0.0  & 0.0$\pm$0.0  & 44.0$\pm$8.6  & \underline{81.3$\pm$6.8} & -- & \textbf{91.0$\pm$0.8} \\
Close Drawer& 85.7$\pm$1.2 & 55.0$\pm$5.7 & 68.0$\pm$0.0 & \textbf{100.0$\pm$0.0} & \underline{97.3$\pm$1.9} & \textbf{100.0$\pm$0.0} & \textbf{100.0$\pm$0.0} \\
\midrule
Mean       & 23.6 & 28.8 & 21.6 & 42.9 & 55.6 & \underline{64.0} & \textbf{73.0} \\
\bottomrule
\end{tabular}
\end{table*}

\subsection{Comparison with the State of the Art}
Table~\ref{tab:main} compares \method{} against representative imitation-learning and mobile manipulation policies. \method{} attains the best mean success rate (73.0\%), leads on five of the seven subtasks, and never collapses anywhere. Every baseline, by contrast, drops to single-digit success on at least one subtask. The margins are substantial over the strongest reported baseline AnchorVLA~\citep{lim2026anchorvla}, which evaluates only six of the seven subtasks, and $+17.4$ mean points over AC-DiT~\citep{chen2025acdit}. This is achieved with a strictly leaner recipe: no 3D point-cloud stream (on which DP3~\citep{ze2024dp3} also relies, to little avail), no two-stage mobility-pretraining curriculum, and a comparable demonstration budget. The spatio-temporal consistency that other pipelines build by hand through point clouds, auxiliary decoders, and privileged GT segmentation~\citep{tu2026sgvla} is inherited in \method{} from video pretraining and sharpened by \cof{}. Two-dimensional observations alone turn out to support this capability surprisingly well.

\begin{table}[t]
\centering
\small
\setlength{\tabcolsep}{10pt}
\renewcommand{\arraystretch}{1.15}
\caption{\textbf{Component ablation.} Starting from the plain WAM, adding \cof{} and then the mobile MoE monotonically improves the mean success rate. Same protocol as Table~\ref{tab:main}. \textbf{Bold} marks the best and \underline{underline} the second best per row.}
\label{tab:components}
\begin{tabular}{lccc}
\toprule
\textbf{Subtask} & \textbf{WAM only} & \textbf{\,+\,\cof{}} & \textbf{\,+\,mobile MoE} \\
\midrule
Pick Apple   & 39.7$\pm$2.1 & \underline{44.7$\pm$1.5} & \textbf{46.0$\pm$0.8} \\
Place Apple  & 45.7$\pm$1.5 & \underline{52.3$\pm$2.5} & \textbf{63.7$\pm$3.2} \\
Open Fridge  & 97.0$\pm$1.0 & \underline{98.0$\pm$1.0} & \textbf{99.3$\pm$0.5} \\
Pick Bowl    & 39.7$\pm$2.1 & \underline{44.3$\pm$0.6} & \textbf{46.0$\pm$2.6} \\
Place Bowl   & 46.3$\pm$1.5 & \underline{56.3$\pm$1.5} & \textbf{64.7$\pm$1.2} \\
Open Drawer  & \underline{89.7$\pm$2.1} & 87.0$\pm$1.0 & \textbf{91.0$\pm$0.8} \\
Close Drawer & \textbf{100.0$\pm$0.0} & \textbf{100.0$\pm$0.0} & \textbf{100.0$\pm$0.0} \\
\midrule
Mean        & 65.4 & \underline{68.9} & \textbf{73.0} \\
\bottomrule
\end{tabular}
\end{table}

\subsection{Ablation Studies}

\paragraph{Does each component pay its way?}
Table~\ref{tab:components} dissects the full model. Adding \cof{} to the plain WAM lifts the mean success rate from 65.4\% to 68.9\% ($+3.5$), with zero inference cost. Gains appear on both pick and place under a moving viewpoint (Pick Apple $+5.0$, Place Bowl $+10.0$). Stacking the mobile MoE on top adds another $+4.1$, with the largest jump on Place Apple ($+11.4$). That subtask interleaves base repositioning with precise release, the very regime conflict the MoE is built to resolve. Together the two modules compound rather than substitute: foresight densifies temporal supervision while MoE resolves action-space interference.

\begin{table}[t]
\centering
\small
\setlength{\tabcolsep}{6pt}
\caption{\textbf{Parallel vs.\ serial foresight} (reduced 5{,}000-step budget; decoupled action denoising). Dense future supervision helps only when it is chained causally and the chaining module is expressive enough. \textbf{Bold} marks the best and \underline{underline} the second best per row.}
\label{tab:variants}
\begin{tabular}{lcccc}
\toprule
Subtask & WAM only & \;+\,parallel foresight & \;+\,MLP-style \cof{} & \;+\,Transformer-style \cof{} \\
\midrule
Pick Apple   & \underline{24.3$\pm$1.5} & 12.7$\pm$1.2 &  8.3$\pm$0.6 & \textbf{32.3$\pm$1.5} \\
Place Apple  & 21.7$\pm$2.1 & \textbf{36.0$\pm$2.0} & \underline{32.7$\pm$1.2} & 27.3$\pm$1.2 \\
Open Fridge  & \textbf{100.0$\pm$0.0} & \textbf{100.0$\pm$0.0} & \underline{97.0$\pm$1.0} & \textbf{100.0$\pm$0.0} \\
Pick Bowl    & \underline{19.7$\pm$2.1} & 12.7$\pm$1.2 & 12.7$\pm$1.2 & \textbf{36.0$\pm$2.0} \\
Place Bowl   & 34.3$\pm$1.2 & \textbf{56.7$\pm$1.2} & 33.0$\pm$1.0 & \underline{35.7$\pm$2.1} \\
Open Drawer  & \underline{51.7$\pm$0.6} & 47.7$\pm$0.6 & 40.7$\pm$1.5 & \textbf{76.0$\pm$2.0} \\
Close Drawer & \textbf{100.0$\pm$0.0} & \textbf{100.0$\pm$0.0} & \textbf{100.0$\pm$0.0} & \textbf{100.0$\pm$0.0} \\
\midrule
Mean        & 50.2 & \underline{52.3} & 46.3 & \textbf{58.2} \\
\bottomrule
\end{tabular}
\end{table}

\paragraph{Parallel or serial? MLP or transformer?}
\cof{} was not designed in one stroke; Table~\ref{tab:variants} retraces the evolution. The first instinct for densifying supervision is \emph{parallel} foresight: extend the prediction horizon and supervise all future chunks side by side, with no chaining. This helps ($50.2\rightarrow52.3$), confirming that more future supervision is more signal. The gain is modest, however, because parallel targets share no causal structure; the model can fit each future independently without learning how one begets the next. The obvious fix is to serialize: let each foresight step condition on its predecessor's belief. Doing so with the simplest recurrence, an MLP, backfires badly (46.3): the bottleneck is too crude, the chained representation collapses, and the corrupted belief pollutes the shared backbone through gradients. A transformer-based foresight module restores the expressiveness the chain needs and delivers the best result (58.2). The lesson is crisp: \emph{causally chained} supervision beats parallel supervision, but only if the belief state is rich enough to carry a world forward. In short, denser futures help only when the inductive bias matches the causal structure of mobile manipulation.

\begin{table}[t]
\centering
\small
\caption{\textbf{Which layers feed the chain?} Mean S.R.\ (\%) under the reduced budget, decoupled action denoising.}
\label{tab:layers}
\begin{tabular}{lc}
\toprule
Layer selection & Mean S.R. \\
\midrule
All 30 layers               & 37.1 \\
Uniform 20 layers           & 42.7 \\
Uniform 12 layers           & 50.8 \\
\textbf{Uniform 4 layers}   & \textbf{58.2} \\
First and last layer        & 55.6 \\
Last 4 layers               & 52.7 \\
Middle 4 layers             & 50.0 \\
First 4 layers              & 54.7 \\
\bottomrule
\end{tabular}
\end{table}

\begin{table}[t]
\centering
\small
\caption{\textbf{Foresight chain length $K$.} Mean S.R.\ (\%) under the reduced budget.}
\label{tab:depth}
\begin{tabular}{lcccc}
\toprule
$K$ & 1 & 2 & 3 & 4 \\
\midrule
Mean S.R. & 55.7 & 54.6 & \textbf{58.2} & 56.3 \\
\bottomrule
\end{tabular}
\end{table}

\paragraph{Where should the belief state come from?}
Table~\ref{tab:layers} probes which backbone layers should seed the chain. Uniformly sampling four layers wins decisively; both indiscriminate concatenation (all 30 layers, 37.1) and any single-region selection (first/middle/last four) underperform. We read this as a signal-to-interference trade-off: a sparse, depth-spanning sample covers the geometry-to-semantics spectrum, whereas dumping every layer into the fusion MLP drowns the physical signal and drags 30 layers' worth of gradients into one auxiliary head. Empirically, four taps strike the best balance between coverage and gradient noise.

\paragraph{How far should foresight reach?}
Table~\ref{tab:depth} varies the chain length. Performance is remarkably stable ($K{=}1$ to $4$ within 3.6 points), peaking at $K{=}3$. This roughly matches the horizon over which the benchmark's subtask dynamics remain predictable. We fix $K{=}3$ throughout. Longer chains add little once the near-term causal window is already covered.

\begin{table}[t]
\centering
\small
\setlength{\tabcolsep}{5pt}
\caption{\textbf{One MoE or two action experts?} Mean S.R.\ (\%) under the reduced budget. ``M$\!\rightarrow\!$L'': one-way mask where manipulation tokens attend to locomotion tokens; ``L$\!\rightarrow\!$M'' is the reverse.}
\label{tab:moedesign}
\begin{tabular}{lcccc}
\toprule
 & MoE (ours) & Bi-dir. & M$\rightarrow$L & L$\rightarrow$M \\
\midrule
Mean S.R. & \textbf{58.2} & 48.8 & 46.9 & 44.6 \\
\bottomrule
\end{tabular}
\end{table}

\paragraph{One MoE or two action experts?}
A tempting alternative to our MoE is architectural separation: split the action expert into \emph{two} equal-sized experts, one denoising the locomotion dimensions and one the manipulation dimensions, with all other attention kept identical to ours. Table~\ref{tab:moedesign} evaluates three masks between the two: bidirectional, manipulation-attends-locomotion, and the reverse. Every variant falls well short of our three-expert MoE. The reason, we believe, is that whole-body dimensions are not separable streams but facets of one simultaneous motion. A hard split halves the data each expert effectively learns from, severs the shared parameters that consolidate common structure (kinematics, visual grounding), and forces base--arm coordination to squeeze entirely through cross-expert attention. Restricting that channel one-way degrades results further still. Soft routing over a shared expert keeps one coherent trajectory-level representation while letting specialization emerge: separation of concerns without separation of knowledge. Soft MoE thus outperforms hard architectural splits under the same parameter budget.

\begin{table}[t]
\centering
\small
\caption{\textbf{Inference latency} for one prediction cycle on the same mobile manipulation task (NVIDIA A800).}
\label{tab:latency}
\begin{tabular}{lc}
\toprule
Model & Latency (ms) $\downarrow$ \\
\midrule
Motus~\citep{bi2026motus}        & 4950 \\
LingBot-VA~\citep{li2026lingbotva} & 8126 \\
\textbf{\method{} (ours)}        & \textbf{938} \\
\bottomrule
\end{tabular}
\end{table}

\paragraph{Deployment efficiency.}
Table~\ref{tab:latency} times one full prediction cycle on an NVIDIA A800. Imagine-then-execute WAMs must denoise an entire video before acting. \method{} discards both the foresight chain and the video branch at deployment, encodes the current observation once, and denoises actions against the cached backbone features. This yields $5.3\times$ and $8.7\times$ speedups over Motus and LingBot-VA, respectively. The world model pays for itself at training time and then gets out of the way. Decoupled denoising is therefore not only a modeling choice but a practical deployment requirement for mobile platforms.

\begin{table}[!htb]
\centering
\small
\setlength{\tabcolsep}{4pt}
\caption{\textbf{Real-robot success rates} on the ARX Lift2 across five tasks of increasing horizon ($T_a\rightarrow T_e$).}
\label{tab:real}
\begin{tabular}{lccccc}
\toprule
Method & $T_a$ & $T_b$ & $T_c$ & $T_d$ & $T_e$ \\
\midrule
$\pi_{0.5}$~\citep{black2025pi05} & 35\% & 25\% & 10\% & 10\% & 0\% \\
\textbf{\method{} (ours)} & \textbf{55\%} & \textbf{35\%} & \textbf{25\%} & \textbf{20\%} & \textbf{15\%} \\
\bottomrule
\end{tabular}
\end{table}

\subsection{Real-Robot Experiments}
We fine-tune \method{} on teleoperated demonstrations collected with an ARX Lift2 mobile manipulator and evaluate five household tasks of increasing horizon, from single-step drawer opening ($T_a$), through shelf and drawer pick-and-place variants ($T_b$--$T_d$), to a compound open--fetch--deposit--close sequence ($T_e$) (task overview in Figure~\ref{fig:teaser}, top-left; full setup in the Appendix). The suite deliberately stresses viewpoint change, base--arm coordination, and multi-stage causal chaining under the same RGB-only observation interface used in simulation. Table~\ref{tab:real} compares against $\pi_{0.5}$~\citep{black2025pi05} fine-tuned on the same data. \method{} wins on every task. The margin is most telling on the longest-horizon $T_e$: $\pi_{0.5}$ never succeeds while \method{} reaches 15\%. This ordering confirms our central claim that the value of causally chained representations grows with the causal depth of the task.

\begin{figure}[t]
\centering
\IfFileExists{Figures/figure3.pdf}%
{\includegraphics[width=0.98\columnwidth]{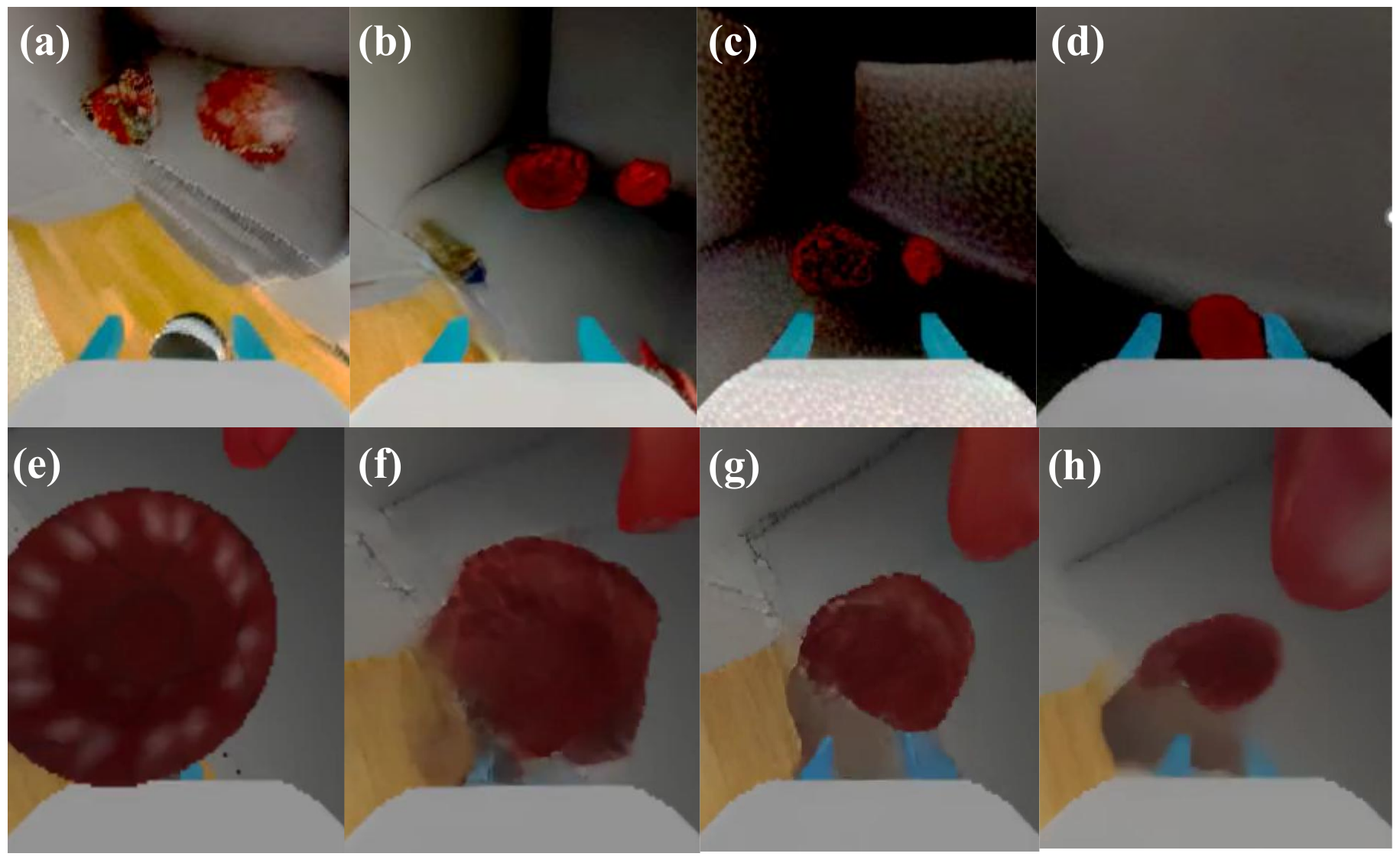}}%
{\fbox{\parbox[c][5.5cm][c]{0.92\columnwidth}{\centering \emph{Placeholder for \texttt{Figures/figure3.pdf}: (a)--(d) futures predicted by our full model during fridge picking; (e)--(h) predictions of the merely fine-tuned Wan backbone.}}}}
\caption{\textbf{Generated futures during fridge picking.} (a)--(d): our full model (\cof{}\,+\,mobile MoE); (e)--(h): the Wan backbone merely fine-tuned on the same data. Ours preserves perspective and shape consistency under base motion; the baseline distorts both.}
\label{fig:qualitative}
\end{figure}

\subsection{Qualitative Analysis of Generated Futures}
Although video generation is disabled at deployment, it remains a window into what the model has learned under \cof{} and the mobile MoE. Figure~\ref{fig:qualitative}(a)--(d) shows the futures predicted by our full model while executing the fridge-picking task; Figure~\ref{fig:qualitative}(e)--(h) shows the same rollout predicted by the Wan backbone merely fine-tuned on the dataset. The contrast is stark. Our predictions respect the rules of spatial perception: as the base advances, objects scale consistently with distance and retain their silhouettes. The fine-tuned-only baseline violates the near-large--far-small law of perspective: objects fail to grow as the robot approaches, and their shapes warp across frames. This corrupts precisely the visual evolution from which coherent actions must be inferred, and helps explain why denser foresight supervision improves closed-loop success even though video is discarded at test time.

\subsection{Failure Analysis and Limitations}
Failures concentrate on pick-and-place subtasks. Categorizing all failed episodes: \emph{localization errors} (grasp or release pose misses the tolerance) account for ${\sim}40\%$; \emph{collision interference} ${\sim}25\%$; \emph{no recovery after an initial failure} ${\sim}20\%$; \emph{failure to restore the rest pose after completion} ${\sim}10\%$; and \emph{premature drops} ${\sim}5\%$. The dominant mode clusters on target poses at the fringe of the robot's workspace, where success demands near-exact inverse kinematics and slight perturbations render the commanded pose unreachable. Open and close subtasks fail far less often: their geometry is coarser and less sensitive to centimeter-level endpoint error. These modes chart the road ahead: stronger spatio-temporal consistency, finer collision awareness, and closed-loop error recovery.

\FloatBarrier  
\section{Conclusion}
We presented \method{}, an early bridge between WAM and mobile manipulation. \method{} fuses a pretrained video diffusion transformer with a lightweight action expert via layerwise joint attention, specializes the action pathway with a three-expert (shared, locomotion, manipulation) mixture, and densifies temporal supervision with Chain-of-Foresight, an RNN-style latent dynamics chain that costs nothing at inference. Together, these convert internet-scale motion priors into whole-body control. \method{} sets a new state of the art on ManiSkill-HAB with RGB-only inputs and single-stage training, runs at policy-level latency ($5$--$8\times$ faster than imagine-then-execute WAMs), and transfers to a real ARX Lift2 with advantages that grow with task horizon.

Looking forward, we see three natural extensions: post-training with reinforcement or interactive corrections, richer geometric grounding beyond RGB, and applying \cof{} during large-scale WAM pretraining to compound its benefits. We hope \method{} persuades the community that mobile manipulation is not merely a harder benchmark for world action models, but their most natural home.

\FloatBarrier  
\bibliographystyle{assets/plainnat}
\bibliography{mobilewam}

\clearpage
\FloatBarrier
\appendix
\setcounter{secnumdepth}{2}
\setcounter{figure}{0}
\setcounter{table}{0}
\renewcommand{\thefigure}{S\arabic{figure}}
\renewcommand{\thetable}{S\arabic{table}}

\beginappendix
\vspace{0.5em}

\section{Implementation Details}
\paragraph{Architecture.}
The world expert is a 30-block video diffusion transformer (hidden width 3072, feed-forward width 14336, 24 attention heads of dimension 128) initialized from a pretrained open-source text-and-image-to-video model; its causal 3D VAE (48 latent channels, $16\times$ spatial and $4\times$ temporal compression) and T5-family text encoder are frozen throughout. The action expert has 30 blocks with hidden width 1024, feed-forward width 4096, and 24 heads of dimension 128, so per-head dimensions match the backbone and joint attention concatenates heads directly. Action chunks ($H{=}4$, $d_a{=}13$) are embedded by a linear layer; the diffusion timestep enters through an AdaLN-style modulation. The proprioceptive state ($d_s{=}18$: goal position, grasp indicator, object pose, and end-effector pose in the base frame) is linearly projected and appended to the text context.

\paragraph{Chain-of-Foresight.}
The fusion MLP has two layers. Each depth module $F_k$ comprises three transformer blocks mirroring the backbone block design (self-attention with rotary position embeddings under a block-causal mask, cross-attention to language/proprioception/current-observation tokens, feed-forward), operating at the backbone width. All $F_k$ have independent weights initialized cyclically from the final backbone blocks. Foresight targets use a more aggressive noise-schedule shift (10 vs.\ 5 for the main branches), reflecting the higher uncertainty of deeper futures. Gradients from the chain flow back into the four tapped backbone layers through the fusion MLP; no stop-gradient is applied along the chain.

\paragraph{Mobile MoE.}
Each action-expert feed-forward layer is replaced by three experts (shared, locomotion, manipulation), each a clone of the pretrained dense layer at initialization. The router is a zero-initialized linear layer on the mean-pooled noisy-action embedding with softmax temperature 1.0; outputs are convex combinations of the three experts. No auxiliary balancing loss is used.

\paragraph{Optimization.}
AdamW ($\beta_1{=}0.9$, $\beta_2{=}0.95$, weight decay 0.01), learning rate $10^{-5}$ with cosine decay and 5\% linear warmup, gradient clipping at 1.0, bf16 mixed precision with ZeRO-style sharding. The main models train with a global batch size of 256; reduced-budget ablations train for 5{,}000 optimizer steps. Loss weights are $\lambda_{\mathrm{v}}{=}\lambda_{\mathrm{a}}{=}1$ and $\lambda{=}0.1$ for \cof{}, with depth decay $\bs{w}=(0.4, 0.2, 0.1)$ and $K{=}3$. Text and proprioception conditioning are dropped with probability 0.1 during training for classifier-free-guidance-style robustness; images receive color jitter and mild corruption augmentation with probability 0.5.

\paragraph{Decoupled inference.}
At deployment, the current composited observation is encoded by the frozen VAE; a single backbone forward pass over the clean current-observation tokens populates a per-layer key-value cache; the action chunk is then iteratively denoised for 20 flow-matching steps attending to this cache. The foresight chain and future-video tokens are never instantiated. The robot executes all $H{=}4$ actions of a chunk before re-observing and replanning.

\section{Benchmark and Data Details}
The seven SetTable combinations pair four subtask types (pick, place, open, close) with articulated fixtures (fridge, kitchen counter) and target objects (apple, bowl). Episodes randomize object placements and robot spawn poses; the robot must coordinate base motion, torso lift, head gaze, and arm control from RGB observations. Demonstrations are generated by the benchmark's reinforcement-learning-plus-filtering pipeline, which retains trajectories satisfying behavior and safety predicates; we use 1{,}000 training and 100 validation trajectories per combination. Success rates are computed by closed-loop rollout with the benchmark's official completion predicates; we report mean and standard deviation over three independent evaluation runs.

\begin{figure*}[t]
\centering
\IfFileExists{Figures/figureS1-1.pdf}%
  {\includegraphics[width=0.96\textwidth]{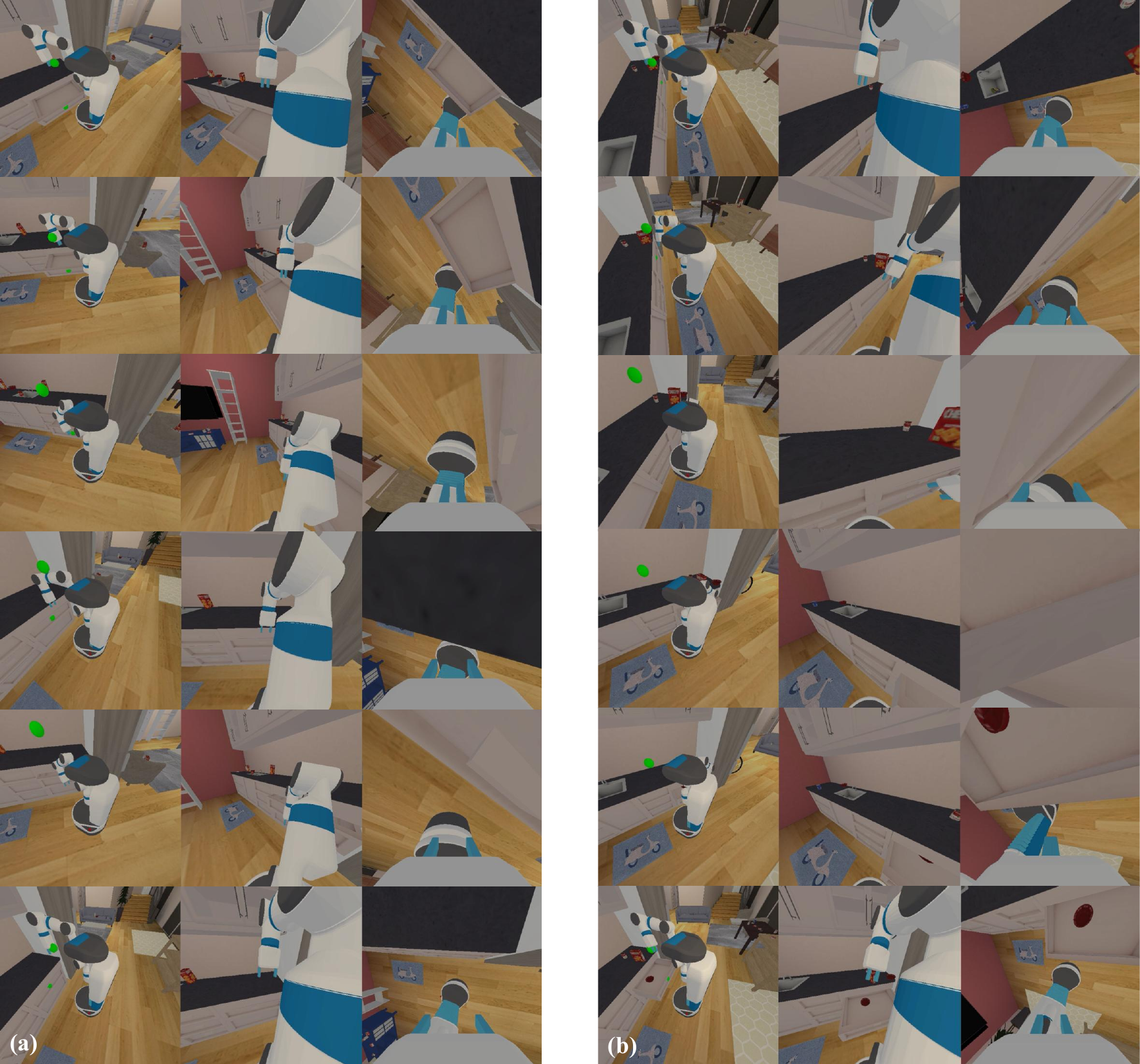}}%
  {\fbox{\parbox[c][8cm][c]{0.92\textwidth}{\centering\small\emph{Placeholder: \texttt{figureS1-1.pdf}.
  Each column is one task (time top$\to$bottom; views left$\to$right: third-person, head, wrist).}}}}
\caption{\textbf{Simulation task suite --- Close/Open Drawer (ManiSkill-HAB).}
\textbf{(a)} Close Drawer: navigate to the kitchen counter and push its drawer closed.
\textbf{(b)} Open Drawer: navigate to the kitchen counter and pull its drawer open.}
\label{fig:sim1}
\end{figure*}

\begin{figure*}[t]
\centering
\IfFileExists{Figures/figureS1-2.pdf}%
  {\includegraphics[width=0.96\textwidth]{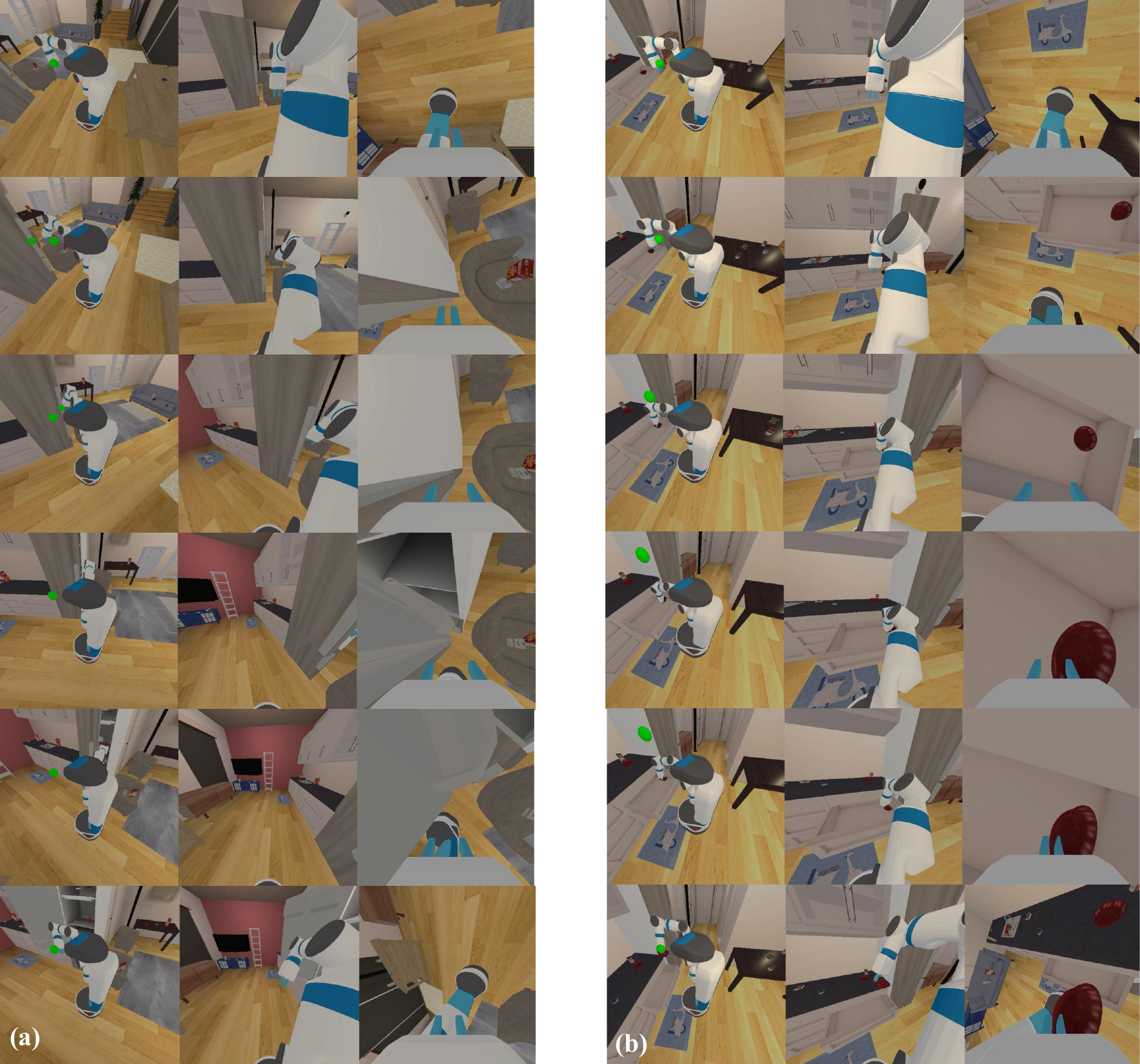}}%
  {\fbox{\parbox[c][8cm][c]{0.92\textwidth}{\centering\small\emph{Placeholder: \texttt{figureS1-2.pdf}.}}}}
\caption{\textbf{Simulation task suite --- Open Fridge Door / Pick Bowl (ManiSkill-HAB).}
\textbf{(a)} Open Fridge Door: approach the refrigerator and pull the door open.
\textbf{(b)} Pick Bowl: grasp the bowl from the kitchen counter top.}
\label{fig:sim2}
\end{figure*}

\begin{figure*}[t]
\centering
\IfFileExists{Figures/figureS1-3.pdf}%
  {\includegraphics[width=0.96\textwidth]{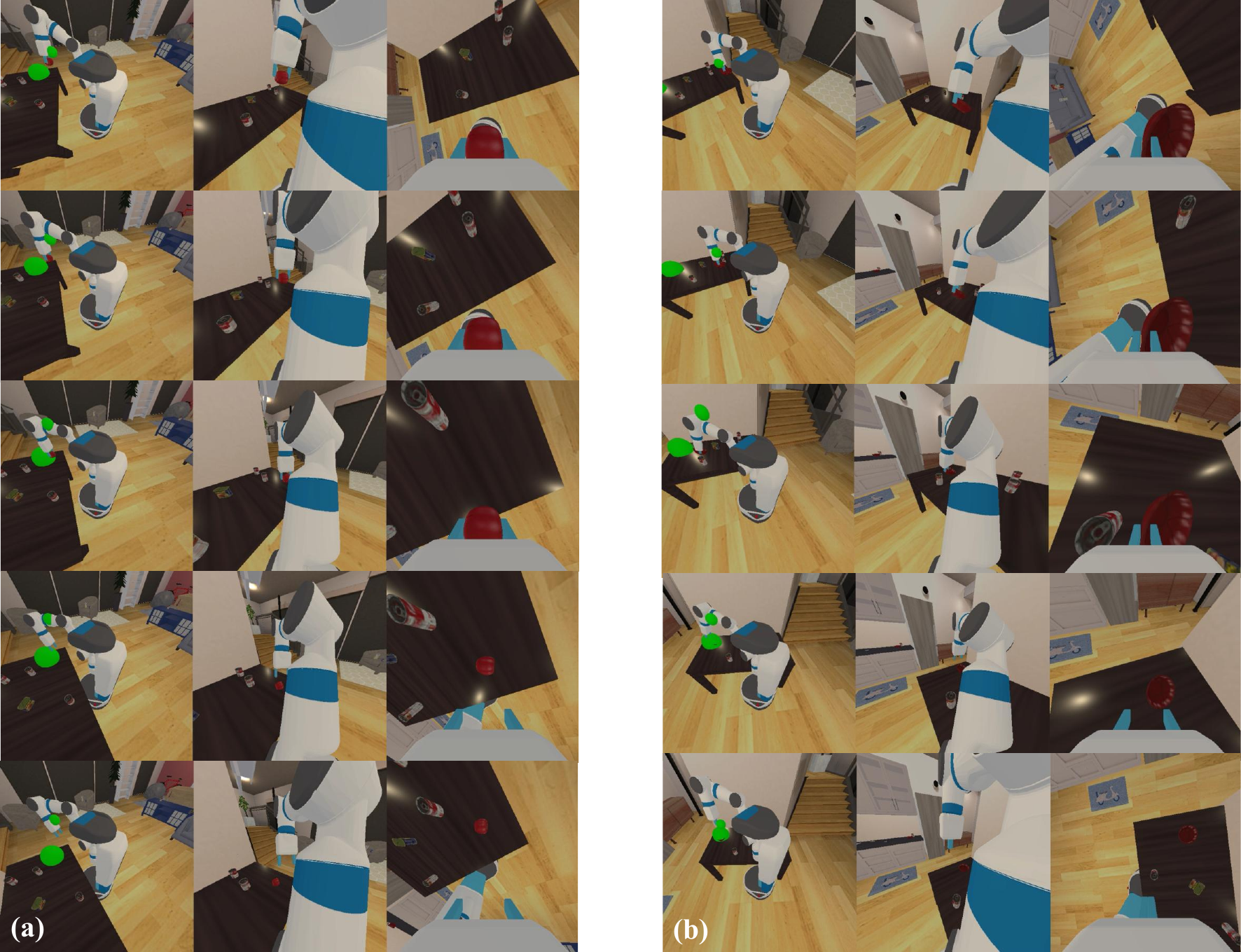}}%
  {\fbox{\parbox[c][8cm][c]{0.92\textwidth}{\centering\small\emph{Placeholder: \texttt{figureS1-3.pdf}.}}}}
\caption{\textbf{Simulation task suite --- Place Apple / Place Bowl (ManiSkill-HAB).}
\textbf{(a)} Place Apple: carry the apple and place it at the target position on the dining table.
\textbf{(b)} Place Bowl: carry the bowl and place it at the target position on the dining table.}
\label{fig:sim3}
\end{figure*}

\begin{figure*}[t]
\centering
\IfFileExists{Figures/figureS1-4.pdf}%
  {\includegraphics[width=0.96\textwidth]{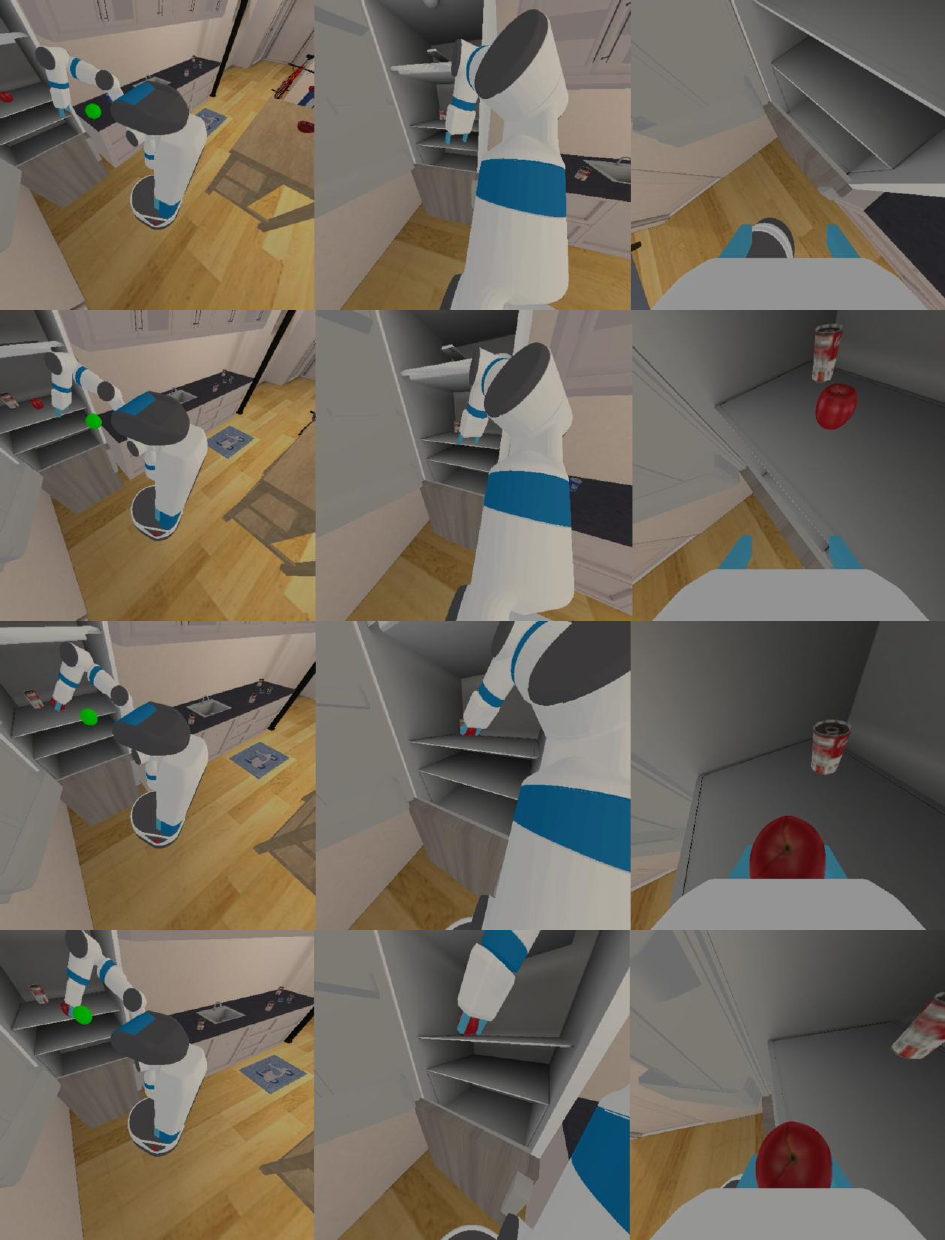}}%
  {\fbox{\parbox[c][8cm][c]{0.92\textwidth}{\centering\small\emph{Placeholder: \texttt{figureS1-4.pdf}.}}}}
\caption{\textbf{Simulation task suite --- Pick Apple (ManiSkill-HAB).}
Pick Apple: navigate to and open the refrigerator door (if not already open), then grasp the apple inside.}
\label{fig:sim4}
\end{figure*}

The seven subtasks are as follows; representative rollout strips are shown in Figures~\ref{fig:sim1}--\ref{fig:sim4}.
\begin{enumerate}
  \item \emph{Close Drawer}: navigate to the kitchen counter and push its drawer closed.
  \item \emph{Open Drawer}: navigate to the kitchen counter and pull its drawer open.
  \item \emph{Open Fridge Door}: approach the refrigerator and pull the door open.
  \item \emph{Pick Bowl}: grasp the bowl from the kitchen counter top.
  \item \emph{Place Apple}: carry the apple and place it at the target position on the dining table.
  \item \emph{Place Bowl}: carry the bowl and place it at the target position on the dining table.
  \item \emph{Pick Apple}: open (or navigate through the open) fridge door and grasp the apple inside the refrigerator.
\end{enumerate}

Each rollout strip in Figures~\ref{fig:sim1}--\ref{fig:sim4} places one task per column with time top to bottom; within each row the three panels are (left to right) the third-person, head, and wrist views.
Full closed-loop rollouts for these subtasks are included in the accompanying supplementary video.

\section{Real-Robot System and Task Suite}
\paragraph{Platform.}
The ARX Lift2 is a wheeled mobile manipulator with a lifting torso, equipped with a head-mounted Intel RealSense 405 RGB camera and a wrist-mounted camera matching our two-view observation format. Because the on-board compute of ARX Lift2 is insufficient to run our ${\approx}$6.5B-parameter model in real time, we deploy \method{} on a remote server equipped with two NVIDIA A800 GPUs (80\,GB each). The robot communicates with the server over Wi-Fi: at each replanning step, the current head and wrist camera images are streamed to the server, the server runs one inference cycle (${\approx}938$\,ms), and the resulting action chunk is transmitted back to the robot for execution. For each task we collect teleoperated demonstrations covering randomized object placements and robot start poses, and fine-tune both \method{} and the $\pi_{0.5}$ baseline on identical data. Evaluation episodes randomize initial conditions within the training distribution; an episode succeeds only if the full task predicate is satisfied.

\begin{figure*}[t]
\centering
\IfFileExists{Figures/figureS2.pdf}%
{\includegraphics[width=0.96\textwidth]{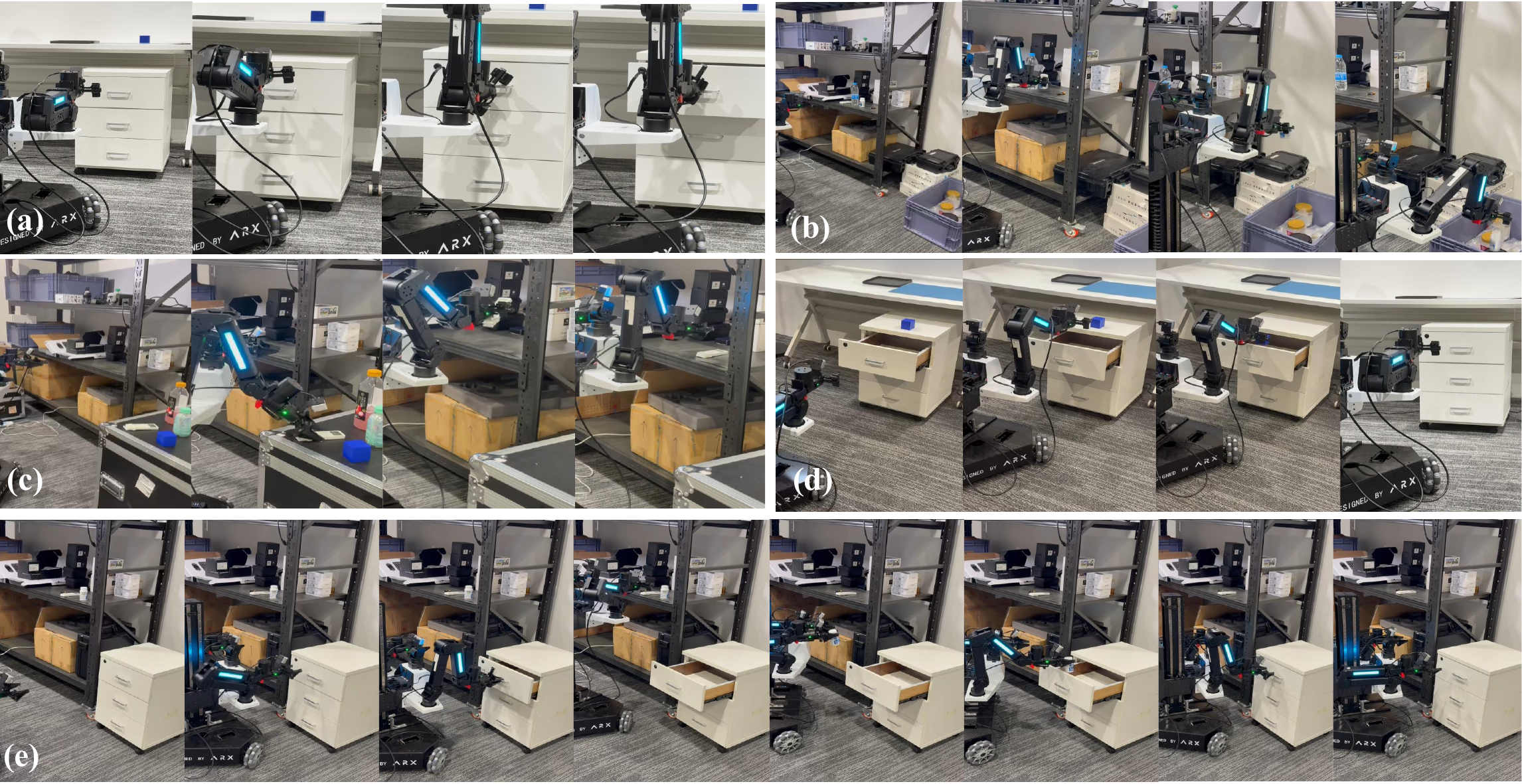}}%
{\fbox{\parbox[c][7cm][c]{0.92\textwidth}{\centering \emph{Placeholder for \texttt{Figures/figureS2.pdf}: representative rollout strips for $T_a$--$T_e$ on the ARX Lift2.}}}}
\caption{\textbf{Real-robot task suite on the ARX Lift2.}
Representative rollout strips for all five tasks $T_a$--$T_e$, ordered by increasing horizon.
\textbf{(a)} $T_a$: Open Drawer.
\textbf{(b)} $T_b$: Shelf to Target.
\textbf{(c)} $T_c$: Pick to Shelf.
\textbf{(d)} $T_d$: Deposit and Close.
\textbf{(e)} $T_e$: Compound sequence.}
\label{fig:realtasks}
\end{figure*}

\paragraph{Task suite.}
The five tasks, ordered by increasing horizon and illustrated in Figure~\ref{fig:realtasks}, are:
\begin{itemize}
    \item $T_a$ (\emph{Open Drawer}): approach a cabinet and pull its drawer open.
    \item $T_b$ (\emph{Shelf to Target}): fetch an item from a shelf and place it at a designated location.
    \item $T_c$ (\emph{Pick to Shelf}): pick an item and place it onto the shelf.
    \item $T_d$ (\emph{Deposit and Close}): put an item into a drawer, then close the drawer.
    \item $T_e$ (\emph{Compound}): open the drawer, fetch an item from the shelf, deposit it into the drawer, and close the drawer---a long-horizon composition of $T_a$--$T_d$ whose later stages causally depend on the earlier ones.
\end{itemize}
Representative real-robot executions of $T_a$--$T_e$ can also be viewed in the accompanying supplementary video.

\section{Additional Qualitative Results}

\begin{figure*}[t]
\centering
\IfFileExists{Figures/figureS3-1.pdf}%
{\includegraphics[width=0.96\textwidth]{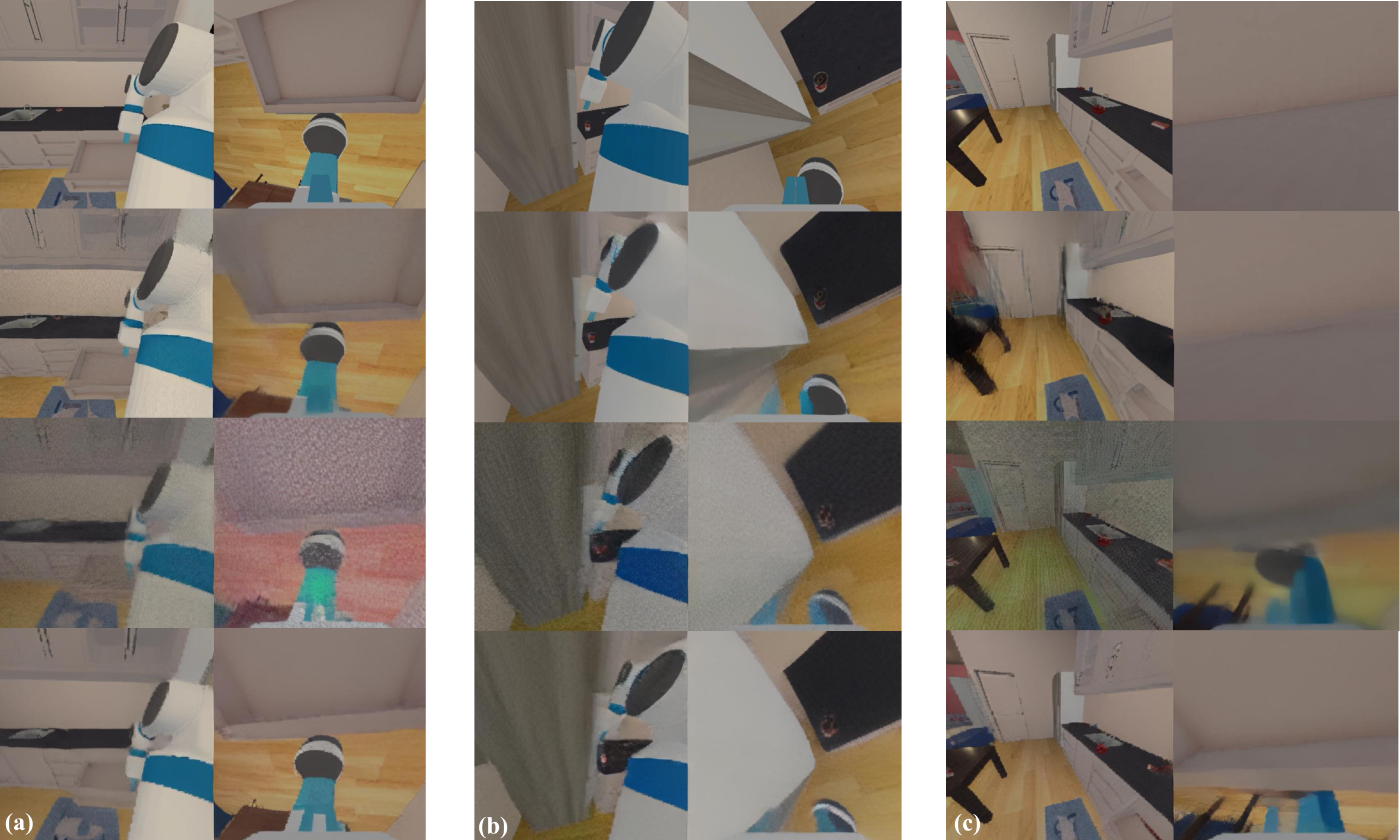}}%
{\fbox{\parbox[c][7cm][c]{0.92\textwidth}{\centering \emph{Placeholder for \texttt{Figures/figureS3-1.pdf}: generated future frames in simulation for three subtasks.}}}}
\caption{\textbf{Video generation in simulation (ManiSkill-HAB).}
Each group shows a partial action segment for one subtask; within each group the left column is the head camera view and the right column is the wrist camera view; the topmost frame of each column is the clean current observation.
\textbf{(a)} Open Drawer.
\textbf{(b)} Close Fridge Door.
\textbf{(c)} Close Drawer.}
\label{fig:qual_sim}
\end{figure*}

\begin{figure*}[t]
\centering
\IfFileExists{Figures/figureS3-2.pdf}%
{\includegraphics[width=0.96\textwidth]{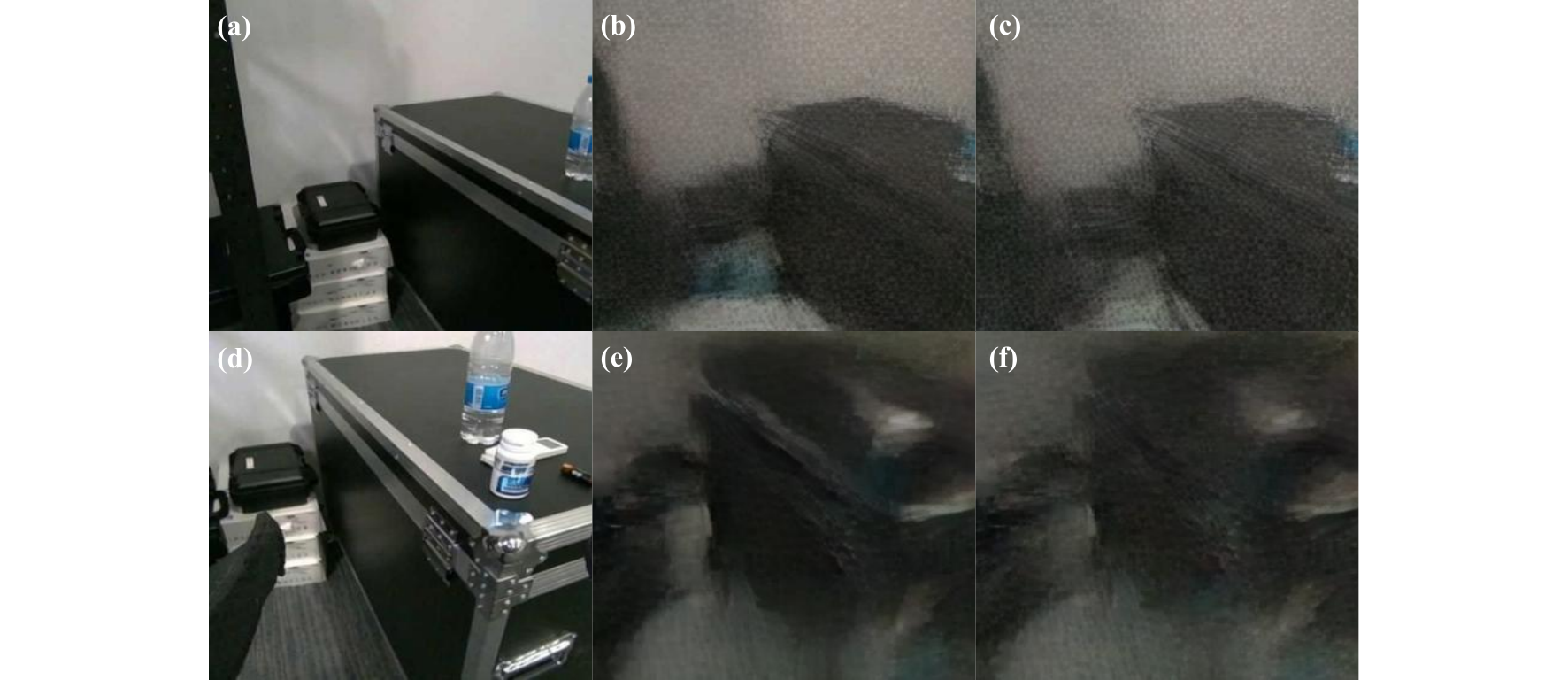}}%
{\fbox{\parbox[c][7cm][c]{0.92\textwidth}{\centering \emph{Placeholder for \texttt{Figures/figureS3-2.pdf}: generated future frames on the real ARX Lift2 during a pick task.}}}}
\caption{\textbf{Video generation on the real ARX Lift2 (pick task).}
A partial action segment during grasping an object from a box.
The leftmost frame in each row is the clean current observation.
\textbf{(a)--(c)} Head camera views at successive timesteps.
\textbf{(d)--(f)} Corresponding wrist camera views.}
\label{fig:qual_real}
\end{figure*}

We visualize video predictions from \method{} during closed-loop execution in ManiSkill-HAB (Figure~\ref{fig:qual_sim}) and on the ARX Lift2 (Figure~\ref{fig:qual_real}). In both settings the predicted future frames capture the intended motion trend: object trajectories, end-effector approach directions, and scene-layout changes. This provides a supervisory signal reflected in the learned action distribution even though video generation is discarded at deployment.

\section{Mobile MoE Expert Routing Analysis}

\begin{figure*}[t]
\centering
\IfFileExists{Figures/figureS8.pdf}%
  {\includegraphics[width=0.97\textwidth]{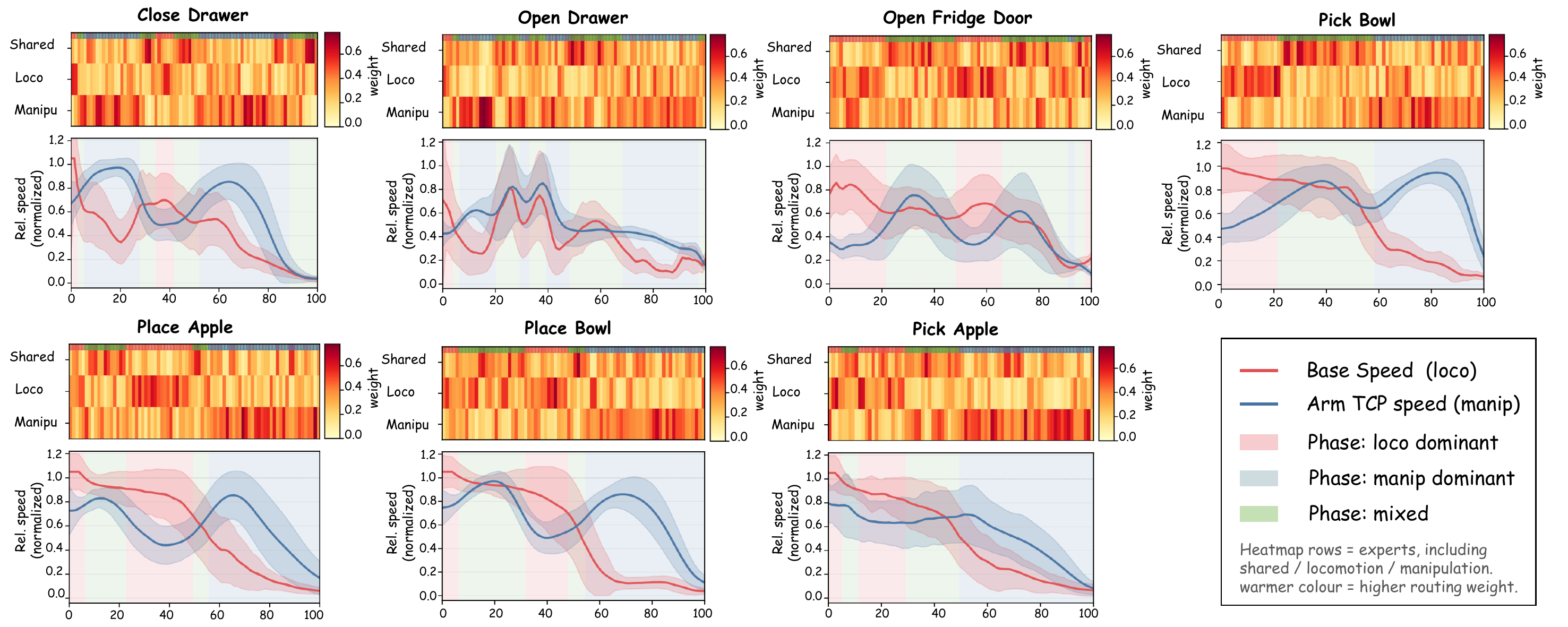}}%
  {\fbox{\parbox[c][8cm][c]{0.92\textwidth}{\centering\small\emph{Placeholder: \texttt{figureS8.pdf} --- MoE routing heatmaps for all 7 SetTable subtasks.}}}}
\caption{\textbf{Mobile MoE expert routing weights aligned with locomotion/manipulation speed (SetTable, 7 subtasks).}
Each panel corresponds to one subtask.
\emph{Top}: Routing-weight heatmap; rows are the three experts (shared / loco / manip, top to bottom), and warmer colours indicate higher soft-routing weight at that timestep.
The colour strip above the heatmap and the matching background shading on the speed plot mark the inferred dominant phase: \textcolor{red}{red} = locomotion-dominant, \textcolor{blue}{blue} = manipulation-dominant, \textcolor{green!60!black}{green} = mixed.
\emph{Bottom}: Mean normalized base speed (red) and arm TCP speed (blue) $\pm$1 std across 1,000 demonstrations.
Across tasks, the brighter heatmap rows broadly track the phase labels from the speed curves: the locomotion expert rises in loco-dominant segments, the manipulation expert in manip-dominant segments, and the shared expert more often during mixed or transition intervals.}
\label{fig:moe_routing}
\end{figure*}

\paragraph{Setup.}
To understand how the mobile MoE routes action tokens across its three experts (shared, locomotion, manipulation), we visualize per-timestep routing weights alongside locomotion and manipulation speed profiles for all seven SetTable subtasks (Figure~\ref{fig:moe_routing}).
For each subtask we compute the mean base speed (norm of the $xy$-velocity from the wheel-controller action) and the mean arm TCP speed (end-effector displacement relative to the base frame per step), averaged over 1,000 demonstration trajectories.
Each timestep is labelled with its dominant modality: \emph{loco-dominant} when base speed exceeds 1.3$\times$ arm speed, \emph{manip-dominant} in the reverse case, and \emph{mixed} otherwise, shown as the thin colour strip above each heatmap.

\paragraph{Routing analysis.}
Across the seven tasks (Figure~\ref{fig:moe_routing}), brighter heatmap rows broadly track the speed-derived phase labels: the locomotion expert rises in loco-dominant segments, the manipulation expert in manip-dominant segments, and the shared expert more often during mixed or transition intervals.
Per-timestep weights fluctuate substantially, yet remain relatively balanced (dominant expert typically 0.4--0.6), with no expert collapsing to near-zero---consistent with soft routing, where all experts always contribute to the output.

\end{document}